\documentclass[11pt]{article}
\usepackage{authblk} 
\usepackage{eacl2017}
\usepackage{times}
\usepackage{url}
\usepackage{graphicx}
\usepackage{color}
\usepackage{tabularx}
\usepackage{xcolor}
\usepackage{tikz}
\usepackage{booktabs}
\usepackage{amssymb}
\usepackage{amsmath}
\usepackage{enumerate}
\usepackage{xspace}
\usepackage{multirow}
\usepackage{paralist,relsize}
\usepackage[small]{caption}
\usepackage{microtype}
\usepackage{soul}

\eaclfinalcopy
\usepackage{bm}

\renewcommand{\vec}[1]{\bm{#1}}
\newcommand{\vhf}[1]{\vec{h}_{\text{#1}}^{\rightarrow}}
\newcommand{\vhb}[1]{\vec{h}_{\text{#1}}^{\leftarrow}}
\newcommand\softmax{\operatorname{softmax}}
\newcommand{\method}[2][]{\texttt{#2}$_{\text{#1}}$\xspace}

\newcommand{\lstm}{\method{LSTM+CTX+BS+POS}}

\newcommand{\bilstmctxbs}{\method{biLSTM+CTX+BS}}
\newcommand{\bilstmctx}{\method{biLSTM+CTX}} 
\newcommand{\bilstmbs}{\method{biLSTM+BS}} 
\newcommand{\bilstmctxbsps}{\method{biLSTM+CTX+BS+POS}}
\newcommand{\baseline}{\method{baseline}}

\newcommand{\word}[1]{\textit{#1}}

\newcommand{\UNK}[1][]{\texttt{UNK}#1\xspace}

\newcommand{\blank}{\underline{\phantom{word}}}

\newcommand{\recall}{\ensuremath{\mathcal{R}}\xspace}

\newcommand{\tabref}[1]{Table~\ref{#1}\xspace}
\newcommand{\figref}[1]{Figure~\ref{#1}\xspace}
\newcommand{\secref}[1]{Section~\ref{#1}\xspace}

\title{Context-Aware Prediction of Derivational Word-forms}

\author[1]{\bf Ekaterina Vylomova}
\author[2]{\bf Ryan Cotterell}
\author[1]{\bf Timothy Baldwin}
\author[1]{\bf Trevor Cohn}
\affil[1]{Department of Computing and Information Systems, The University of Melbourne}
\affil[2]{Center for Language and Speech Processing, Johns Hopkins University}
\affil[  ]{\tt \{evylomova,ryan.cotterell\}@gmail.com }
\affil[ ]{\tt  \{tbaldwin,tcohn\}@unimelb.edu.au}

\date{}

\begin{document}
\maketitle

\begin{abstract}
Derivational morphology is a fundamental and complex characteristic of language.
In this paper we propose the new task of predicting the derivational form
of a given base-form lemma that is appropriate for a given context.
We present an encoder--decoder style neural network to produce a
derived form character-by-character, based on its corresponding
character-level representation of the base form and the context. 
We demonstrate that our model is able to generate valid context-sensitive 
derivations from known base forms, but is less accurate under a lexicon agnostic setting.
\end{abstract}

\section{Introduction}

Understanding how new words are formed is a fundamental task in
linguistics and language modelling, with significant implications for
tasks with a generation component, such as abstractive summarisation and
machine translation. In this paper we focus on modelling derivational
morphology, to learn, e.g., that the appropriate derivational form of
the verb \word{succeed} is \word{succession} given the context \word{As
  third in the line of \blank \ldots}, but is \word{success} in 
\word{The play was a great \blank}.

English is broadly considered to be a morphologically impoverished
language, and there are certainly many regularities in morphological
patterns, e.g., the common usage of \word{-able} to transform a verb
into an adjective, or \word{-ly} to form an adverb from an
adjective. However there is considerable subtlety in English derivational
morphology, in the form of: (a) idiosyncratic derivations; e.g.\ \word{picturesque} vs.~\word{beautiful} vs.~\word{splendid} as
adjectival forms of the nouns \word{picture}, \word{beauty} and
\word{splendour}, respectively; (b) derivational generation in context,
which requires the automatic determination of the part-of-speech (POS) of the stem and the
likely POS of the word in context, and POS-specific derivational rules;
and (c) multiple derivational forms often exist for a given stem,
and these must be selected between based on the context (e.g.\ \word{success}
and \word{succession} as nominal forms of \word{success}, as seen
above). As such, there are many aspects that affect the choice of
derivational transformation, including morphotactics, phonology,
semantics or even etymological characteristics. Earlier works
\cite{thorndike1941teaching} analysed ambiguity of derivational suffixes
themselves when the same suffix might present different semantics
depending on the base form it is attached to (cf.\ \word{beautiful} vs.\ \word{cupful}). Furthermore, as
\newcite{richardson1977lexical} previously noted, even words with quite
similar semantics and orthography such as \word{horror} and
\word{terror} might have non-overlapping patterns: although we observe
regularity in some common forms, for example, \word{horrify} and
\word{terrify}, and \word{horrible} and \word{terrible}, nothing tells us
why we observe \word{terrorize} and no instances of \word{horrorize}, or
\word{horrid}, but not \word{terrid}.
     
In this paper, we propose the new task of predicting a derived form from
its context and a base form. Our motivation in this research is
primarily linguistic, i.e.\ we measure the degree to which it is
possible to predict particular derivation forms from context. A similar
task has been proposed in the context of studying how children master
derivations \cite{singson2000relation}. In their work, children were
asked to complete a sentence by choosing one of four possible
derivations. Each derivation corresponded either to a noun, verb,
adjective, or adverbial form. \newcite{singson2000relation} showed that
childrens' ability to recognize the correct form correlates with their
reading ability. This observation confirms an earlier idea that
orthographical regularities provide a clearer clues to morphological
transformations comparing to phonological rules
\cite{templeton1980spelling,moskowitz1973status}, especially in
languages such as English where grapheme-phoneme correspondences are
opaque.  For this reason we consider orthographic rather than
phonological representations.

In our approach, we test how well models incorporating distributional 
semantics can capture derivational transformations.  
Deep learning models capable of learning real-valued word embeddings
have been shown to perform well on a range of tasks, from language 
modelling \cite{Mikolov+:2013b} to parsing \cite{dyer2015transition} and
machine translation
\cite{bahdanau2014neural}. 
Recently, these models have also been successfully applied to morphological reinflection tasks
\cite{kann2016single,cotterell-EtAl:2016:SIGMORPHON}.


\section{Derivational Morphology}

Morphology, the linguistic study of the internal structure of words, has
two main goals: (1) to describe the relation between different words in
the lexicon; and (2) to decompose words into {\em morphemes}, the
smallest linguistic units bearing meaning.
Morphology can be divided into two types: {\em inflectional} and {\em
  derivational}. Inflectional morphology is the set of processes
through which the word form outwardly displays syntactic information,
e.g., verb tense. It follows that an inflectional affix typically
neither changes the part-of-speech (POS) nor the semantics of the
word.  For example, the English verb \word{to run} takes various
forms: \word{run}, \word{runs} and \word{ran}, all of
which convey the concept ``moving by foot quickly'', but appear in complementary
syntactic contexts.

Derivation, on the other hand, deals with the formation of new
words that have semantic shifts in meaning (often including
POS) and is tightly intertwined with lexical semantics
\cite{light:1996:ACL}. Consider the example of the English noun
\word{discontentedness}, which is derived from the adjective
\word{discontented}. It is true that both words share a close semantic
relationship, but the transformation is clearly more than a simple
inflectional marking of syntax. Indeed, we can go one step further and
define a chain of words \word{content} $\mapsto$ \word{contented}
$\mapsto$ \word{discontented} $\mapsto$ \word{discontentedness}.

In this work, we deal with the formation of deverbal nouns, i.e., nouns
that are formed from verbs. Common examples of this in English include
agentives (e.g., \word{explain} $\mapsto$ \word{explainer}), gerunds
(e.g., \word{explain} $\mapsto$ \word{explaining}), as well as other
nominalisations (e.g., \word{explain} $\mapsto$
\word{explanation}). Nominalisations have varyingly different meanings
from their base verbs, and a key focus of this study is the prediction
of which form is most appropriate depending on the context, in terms of
syntactic and semantic concordance.  Our model is highly flexible and
easily applicable to other related lexical problems.

\begin{figure}[t]
\centering
  \includegraphics[width=0.5\textwidth]{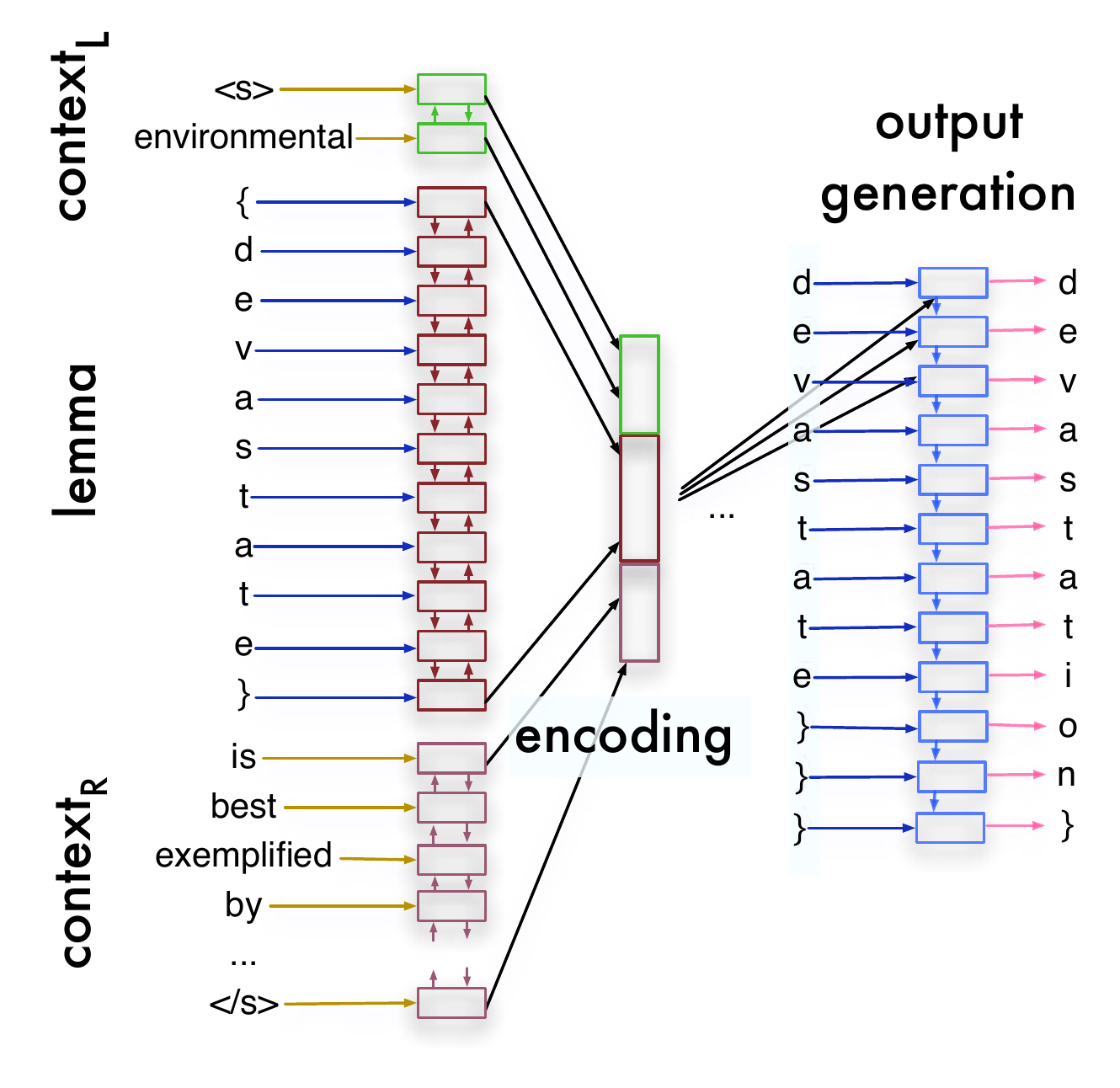}
  \caption{The encoder--decoder model, showing the stem \word{devastate}
    in context producing the form \word{devastation}. Coloured arrows
    indicate shared parameters}
  \label{fig:arch}
\end{figure}

\section{Related Work}

Although in the last few years many neural morphological models have
been proposed, most of them have focused on inflectional morphology
(e.g., see \newcite{cotterell-EtAl:2016:SIGMORPHON}). Focusing on
derivational processes, there are three main directions of research. The
first deals with the evaluation of word embeddings either using a word
analogy task \cite{gladkova2016analogy} or binary relation type
classification \cite{vylomova2015take}. In this context, it has been
shown that, unlike inflectional morphology, most derivational relations
cannot be as easily captured using distributional methods.  Researchers
working on the second type of task attempt to predict derived forms
using the embedding of its corresponding base form and a vector encoding
a ``derivational'' shift. \newcite{guevara2011computing} notes that
derivational affixes can be modelled as a geometrical function over the
vectors of the base forms. On the other hand,
\newcite{lazaridou2013compositional} and
\newcite{DBLP:journals/corr/CotterellS17} represent derivational affixes
as vectors and investigate various functions to combine them with base
forms. \newcite{kisselew2015obtaining} and \newcite{padopredictability}
extend this line of research to model derivational morphology in
German. This work demonstrates that various factors such as part of
speech, semantic regularity and argument structure
\cite{grimshaw1990argument} influence the predictability of a derived
word. The third area of research focuses on the analysis of
derivationally complex forms, which differs from this study in that we
focus on generation.  The goal of this line of work is to produce a
canonicalised segmentation of an input word into its constituent morphs,
e.g., \word{unhappiness}$\mapsto$\word{un}$+$\word{happy}$+$\word{ness}
\cite{cotterell2015labeled,cotterell-vieira-schutze:2016:N16-1}.  Note
that the orthographic change \word{y}$\mapsto$\word{i} has been reversed.



\section{Dataset}

As the starting point for the construction of our dataset, we used the
CELEX English dataset \cite{baayen1993celex}. We extracted
verb--noun lemma pairs from CELEX, covering 24 different
nominalisational suffixes and 1,456 base lemmas. Suffixes only occurring in 5 or fewer lemma
pairs mainly corresponded to loan words and consequently were filtered
out.
We augmented this dataset with verb--verb pairs, one for each verb
present in the verb--noun pairs, to capture the case of a verbal form
being appropriate for the given context.\footnote{We also experimented
  without verb--verb pairs and didn't observe much difference in the
  results.} For each noun and verb lemma, we generated all their
inflections, and searched for sentential contexts of each inflected
token in a pre-tokenised dump of English Wikipedia.\footnote{Based on a
  2008/03/12 dump. Sentences shorter than 3 words or longer than 50
  words were removed from the dataset.}
To dampen the effect of high-frequency words, we applied a heuristic
$\log$ function threshold which is basically a weighted logarithm of the number of the contexts. 
The final dataset contains 3,079 unique lemma pairs represented in 107,041 contextual instances.\footnote{The code and the
  dataset are available at \url{https://github.com/ivri/dmorph}}

\section{Experiments}
In this paper we model derivational morphology as a prediction task,
formulated as follows.  We take sentences containing a derivational form
of a given lemma, then obscure the derivational form by replacing it
with its base form lemma.  The system must then predict the original
(derivational) form, which may make use of the sentential context.
System predictions are judged correct if they exactly match the original
derived form.

\subsection{Baseline}

As a baseline we considered a trigram model with modified Kneser-Ney
smoothing, trained on the training dataset.  Each sentence in the
testing data was augmented with a set of confabulated sentences, where
we replaced a target word with other its derivations or a base form.
Unlike the general task, where we generate word forms as character
sequences, here we use a set of known inflected forms for each lemma
(from the training data).  We then use the language model to score the
collections of test sentences, and selected the variant with the highest
language model score, and evaluate accuracy of selecting the original
word form.

\subsection{Encoder--Decoder Model} \label{sec:encdec}

We propose an encoder--decoder model. The encoder combines the left and
the right contexts as well as a character-level base form
representation:
\begin{align}
 \vec t = 	\max(0, H \cdot [&\vhf{left};\vhb{left};\vhf{right};  \vhb{right}; \nonumber \\
        & \vhf{base};\vhb{base}] + \vec{b_h}), \nonumber
\end{align}
where $\vhf{left}$, $\vhb{left}$, $\vhf{right}$, $\vhb{right}$,
$\vhf{base}$,$\vhb{base}$ correspond to the last hidden states of an
LSTM \cite{hochreiter1997long} over left and right contexts and the character-level representation of the base form (in each case, applied forwards and backwards), respectively;
$H \in \mathbb{R}^{[h \times l \times 1.5,h \times l \times 6]}$ is a
weight matrix, and
$\vec{b}_h \in \mathbb{R}^{[h \times l \times 1.5]}$ is a bias
term. $[;]$ denotes a vector concatenation operation, $h$
is the hidden state dimensionality, and $l$ is the number of layers.

Next we add an extra affine transformation,
$ \vec o = T \cdot \vec t + \vec{b}_o$, where
$T \in \mathbb{R}^{[h \times l \times 1.5,h \times l]}$ and
$\vec{b}_o \in \mathbb{R}^{[h \times l]}$, then $\vec o$ is then fed into the decoder:
\begin{align}
g(&\vec{c}_{j+1}|  \vec{c}_j, \vec o, l_{j+1}) = \nonumber\\ \nonumber
&\softmax ( R \cdot \vec{c}_j +  \max{(B \cdot \vec o, S \cdot \vec{l}_{j+1})}+\vec {b}_d) ,
\end{align}
where $\vec{c}_j$ is an embedding of the $j$-th character of the
derivation, $\vec{l}_{j+1}$ is an embedding of the corresponding base
character, $B, S, R$ are weight matrices, and $\vec {b}_d$ is a bias
term.

We now elaborate on the design choices behind the model architecture
which have been tailored to our task.  We supply the model with the
$l_{j+1}$ character prefix of the base word to enable a copying
mechanism, to bias the model to generate a derived form that is
morphologically-related to the base verb. In most cases, the derived form
is longer than its stem, and accordingly, when we reach the end of the
base form, we continue to input an end-of-word symbol.  We provide the
model with the context vector $\vec o$ at each decoding step. It has
been previously shown \cite{hoang2016incorporating} that this yields
better results than other means of incorporation.\footnote{We tried to
  feed the context information at the initial step only, and this led to
  worse prediction in terms of context-aware suffixes.}  Finally, we use
max pooling to enable the model to switch between copying of a stem or
producing a new character.

\begin{table}[t]
\centering
\footnotesize
\begin{tabular}{lcc}
  \toprule
  & \textbf{Shared} & \textbf{Split}\\
   \midrule
\baseline      & 0.63 & --- \\ 
\bilstmbs & 0.58 & 0.36 \\
\bilstmctx & 0.80 & 0.45 \\
\bilstmctxbs & 0.83 & 0.52 \\
\bilstmctxbsps  & 0.89  & 0.63 \\ 
\lstm    & 0.90 & 0.66 \\ 
\bottomrule
\end{tabular}
\caption{Accuracy for predicted lemmas (bases and derivations) on shared and split lexicons}
\label{tab-mpred}
\end{table}

\subsection{Settings}

We used a 3-layer bidirectional LSTM network, with hidden dimensionality
$h$ for both context and base-form stem states of 100, and character
embedding $\vec{c}_j$ of 100.\footnote{We also experimented with 15
  dimensions, but found this model to perform worse.} We used pre-trained
300-dimensional Google News word embeddings
\cite{Mikolov+:2013b,Mikolov+:2013c}. During the training of the model,
we keep the word embeddings fixed, for greater applicability to unseen
test instances. All tokens that didn't appear in this set were replaced
with \UNK sentinel tokens. The network was trained using SGD with momentum until
convergence.

\subsection{Results}


With the encoder--decoder model, we experimented with the
encoder--decoder as described in \secref{sec:encdec} (
``\bilstmctxbs''), as well as several variations, namely: excluding
context information (``\bilstmbs''), and excluding the bidirectional
stem (``\bilstmctx'').  We also investigated how much improvement we can
get from knowing the POS tag of the derived form, by presenting it
explicitly to the model as extra conditioning context
(``\bilstmctxbsps''). The main motivation for this relates to gerunds,
where without the POS, the model often overgenerates nominalisations. We
then tried a single-directional context representation, by using only
the last hidden states, i.e., $\vhf{left}$ and $\vhb{right}$,
corresponding to the words to the immediate left and right of the
wordform to be predicted (``\lstm'').
 
We ran two experiments: first, a shared lexicon experiment, where every
stem in the test data was present in the training data; and second,
using a split lexicon, where every stem in the test data was
\textit{unseen} in the training data. The results are presented in
\tabref{tab-mpred}, and show that: (1) context has a strong impact on
results, particularly in the shared lexicon case; (2) there is strong
complementarity between the context and character representations,
particularly in the split lexicon case; and (3) POS information is
particularly helpful in the split lexicon case.  Note that most of the
models significantly outperform our baseline under shared lexicon
setting. The baseline model doesn't support the split lexicon setting
(as the derivational forms of interest, by definition, don't occur in
the training data), so we cannot generate results in this setting.


\subsection{Error Analysis}
\begin{figure}[t]
\centering
  \includegraphics[width=1.0\columnwidth]{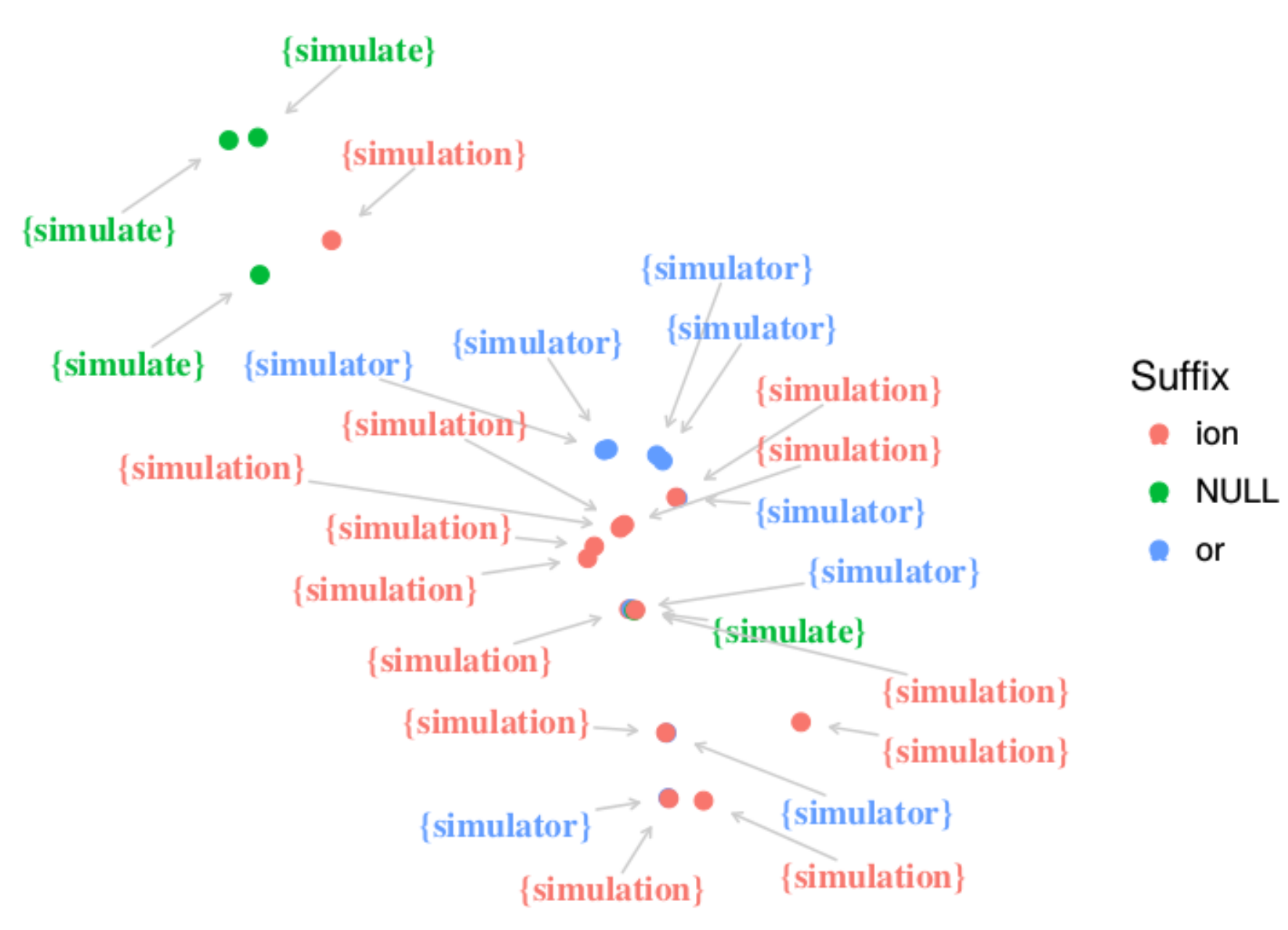}
  \caption{An example of t-SNE projection  \cite{maaten2008visualizing} of context representations for \textit{simulate}}
  \label{fig:tsne}
\end{figure}

We carried out error analysis over the produced forms of the \lstm
model. First, the model sometimes struggles to differentiate between
nominal suffixes: in some cases it puts an agentive suffix (\word{-er}
or \word{-or}) in contexts where a non-agentive nominalisation (e.g.\
\word{-ation} or \word{-ment}) is appropriate. As an illustration of
this, \figref{fig:tsne} is a
t-SNE projection of the context representations for \word{simulate}
vs. \word{simulator} vs.\ \word{simulation}, showing that the different
nominal forms have strong overlap.
Secondly, although the model learns whether to copy or produce a new
symbol well, some forms are spelled incorrectly. Examples of this
are \word{studint}, \word{studion} or even \word{studyant} rather than
\word{student} as the agentive nominalisation of \word{study}.  Here,
the issue is opaqueness in the etymology, with \word{student} being
borrowed from the Old French \word{estudiant}. For
transformations which are native to English, for example,
\word{-ate} $\mapsto$ \word{-ation}, the model is much more accurate. 
\tabref{tab:recall} shows recall values achieved for various suffix types. 
We do not present precision since it could not be reliably estimated 
without extensive manual analysis.  

In the split lexicon setting, the model sometimes misses double consonants at
the end of words, producing \word{wraper} and \word{winer} and is biased towards 
generating mostly productive suffixes. An example of the last case might be \word{stoption} 
in place of \word{stoppage}. We also studied how much the training 
size affects the model's accuracy by reducing the data from 1,000 to  60,000 instances (maintaining a balance over lemmas).
Interestingly, we didn't observe a significant reduction in accuracy. 
Finally, note that under the split lexicon setting, the model is agnostic of existing derivations, sometimes over-generating possible forms. A nice illustration of that is \word{trailation}, 
\word{trailment} and \word{trailer} all being produced in the contexts of \word{trailer}.
In other cases, the model might miss some of the derivations, for instance, predicting only \word{government} in the contexts of \word{governance} and \word{government}. We hypothesize that it is either due to 
very subtle differences in their contexts, or the higher productivity of \word{-ment}.

Finally, we experimented with some nonsense stems, overwriting
sentential instances of \word{transcribe} to generate context-sensitive
derivational forms. \tabref{tab:nons-set} presents the nonsense stems, 
the correct form of \word{transcribe} for a given context, and the
predicted derivational form of the nonsense word.
Note that the base form is used correctly (top row) for three of the
four nonsense words, and that despite the wide variety of output forms,
they resemble plausible words in English. By looking at a larger slice
of the data, we observed some regularities. For instance, \word{fapery}
was mainly produced in the contexts of \word{transcript} whereas
\word{fapication} was more related to
\word{transcription}. \tabref{tab:nons-set} also shows that some of the
stems appear to be more productive than others.

\begin{table}
\centering
\resizebox{\columnwidth}{!}{
  \begin{tabular}{l@{\,\,}cc  l@{\,\,}cc  l@{\,\,}cc  l@{\,\,}c }
    Affix & \recall && Affix & \recall && Affix & \recall && Affix & \recall\\
    \cmidrule{1-2}
    \cmidrule{4-5}
    \cmidrule{7-8}
    \cmidrule{10-11}
    -\word{age}   & .93 && -\word{al}   & .95 && -\word{ance}  & .75 && -\word{ant} & .65 \\
    -\word{ation} & .93 && -\word{ator} & .77 && -\word{ee}    & .52 && -\word{ence} & .82\\
    -\word{ent}   & .65 && -\word{er}   & .87 && -\word{ery}   & .84 && -\word{ion} & .93 \\
    -\word{ist}   & .80 && -\word{ition}& .89 && -\word{ment}  & .90 && -\word{or} & .64 \\
   -\word{th}     & .95 && -\word{ure}  & .77 && -\word{y}     & .83 && NULL & .98\\
  \end{tabular}
}
  \caption{Recall for various suffix types. Here ``NULL'' corresponds to verb--verb cases}
\label{tab:recall}
  \end{table}

\section{Conclusions and Future Work}

We investigated the novel task of context-sensitive derivation
prediction for English, and proposed an encoder--decoder model to
generate nominalisations. Our best model achieved an accuracy of 90\% on
a shared lexicon, and 66\% on a split lexicon. This suggests that there
is regularity in derivational processes and, indeed, in many cases the
context is indicative. As we mentioned earlier, there are still many
open questions which we leave for future work. Further, we plan to scale
to other languages and augment our dataset with Wiktionary data, to
realise much greater coverage and variety of derivational forms.

\begin{table}[t]
\resizebox{\columnwidth}{!}{
\begin{tabular}{llllll}
\toprule
  Original && \multicolumn{4}{c}{Target Lemma}\\
  \cmidrule{1-1}
  \cmidrule{3-6}
  \word{transcribe} && \word{laptify}  &\word{fape} &\word{crimmle} &\word{beteive} \\
\midrule
\word{transcribe}    && \word{laptify} & \word{fape} & \word{crimmle} & \word{beterve}\\
\word{transcription} && \word{laptification} & \word{fapery} & \word{crimmler} & \word{betention}\\
\word{transcription} && \word{laptification} & \word{fapication} & \word{crimmler} & \word{beteption}\\
\word{transcription} && \word{laptification} & \word{fapionment} & \word{crimmler} & \word{betention}\\
\word{transcription} && \word{laptification} & \word{fapist} & \word{crimmler} & \word{betention}\\
\word{transcription} && \word{laptification} & \word{fapist} & \word{crimmler} & \word{beteption}\\
\word{transcript}    && \word{laptification} & \word{fapery} & \word{crimmler} & \word{betention}\\
\word{transcript}    && \word{laptification} & \word{fapist} & \word{crimmler} & \word{beteption}\\
\bottomrule
\end{tabular}}
\caption{An experiment with nonsense ``target'' base forms generated in sentence
  contexts of the ``original'' word \word{transcribe}}
\label{tab:nons-set}
\end{table}

\section{Acknowledgments}
We would like to thank all reviewers for their valuable comments and suggestions. The second author was supported by a DAAD Long-Term Research
Grant and an NDSEG fellowship. This research was supported in part by
the Australian Research Council.

\bibliography{eacl2017}
\bibliographystyle{eacl2017}

\end{document}